\documentclass{article}

    \PassOptionsToPackage{numbers, compress}{natbib}

     \usepackage[preprint]{neurips_2020}

\usepackage[numbers]{natbib}

\usepackage[utf8]{inputenc} 
\usepackage[T1]{fontenc}    
\usepackage{hyperref}       
\usepackage{url}            
\usepackage{booktabs}       
\usepackage{amsfonts}       
\usepackage{nicefrac}       
\usepackage{microtype}      
\usepackage{amsmath}
\usepackage{subcaption}
\usepackage{graphicx}
\usepackage{wrapfig}
\usepackage{epigraph}
\usepackage[export]{adjustbox}
\usepackage{booktabs} 
\usepackage{wrapfig}
\usepackage{color}
\usepackage[inline]{enumitem}

\title{The Scattering Compositional Learner: \\ Discovering Objects, Attributes, Relationships in Analogical Reasoning}

\author{
  Yuhuai Wu\thanks{Equal contribution.} \\
  University of Toronto\\
  Vector Institute\\
  \texttt{ywu@cs.toronto.edu} \\
  \And
  Honghua Dong$^{*}$\\
  University of Toronto\\
  Vector Institute\\
  \texttt{dhh19951@gmail.com } \\ 
  \\
  \AND
  Roger Grosse\\
  University of Toronto\\
  Vector Institute \\
  \texttt{rgrosse@cs.toronto.edu} \\
  \And
  Jimmy Ba\\
  University of Toronto\\
  Vector Institute\\
  \texttt{jba@cs.utoronto.ca} \\
  }

\newif\ifcomments
\commentstrue

\begin{document}

\maketitle
\begin{abstract}


In this work, we focus on an analogical reasoning task that contains rich compositional structures, Raven's Progressive Matrices (RPM). To discover compositional structures of the data, we propose the Scattering Compositional Learner (SCL), an architecture that composes neural networks in a sequence.
Our SCL achieves state-of-the-art performance on two RPM datasets, with a $48.7\%$ relative improvement on Balanced-RAVEN and $26.4\%$ on PGM over the previous state-of-the-art. We additionally show that our model discovers compositional representations of objects' attributes (e.g., shape color, size), and their relationships (e.g., progression, union). 
We also find that the compositional representation makes the SCL significantly more robust to test-time domain shifts and greatly improves zero-shot generalization to previously unseen analogies.


\end{abstract}
\section{Introduction}
\label{sec:intro}
As humans, we routinely use compositions of simpler concepts to make sense of the world.
For instance, consider the mundane feature of everyday language which lets us compose concepts for living things and the activities they are engaged in: ``a running cat'', ``a running elephant'', ``a sleeping cat'', ``a sleeping elephant''.  
By exploiting the compositional structure of concepts, human learners enjoy strong generalization from limited data in various domains~\cite{Lake2018BuildingMT}, making ``infinite use of finite means''~\cite{chomsky1965}. We are capable of comprehending novel situations as compositions of simple and known elements. For example, we can visualize the concept of ``a flying elephant'' despite never having seen one, by combining the familiar notions of ``flying'' and ``an elephant''.

The state-of-the-art deep neural networks have achieved superhuman performance in many vision and language tasks. However, discovering the underlying structure remains a challenge, even for modern deep learning methods ~\cite{ fodor_connectionism_1988, Lake2018GeneralizationWS, Loula2018RearrangingTF, Bahdanau2019SystematicGW, Bahdanau2019CLOSUREAS, Keysers2020MeasuringCG}. Previous studies~\cite{Lake2018GeneralizationWS, Loula2018RearrangingTF} attribute poor generalization to the lack of composition rules in the neural architecture design. In this paper, we introduce a novel neural network architecture to learn a compositional hidden representation. The explicit factorization allows the network to generalize to novel situations.

In this paper, we focus on an analogy completion task that contains rich compositional structures, the Raven's Progressive Matrices (RPM) task~\cite{raven1936performances,John2003}. 
In one of the RPM datasets, RAVEN~\cite{zhang2019raven}, one is given a $3\times 3$ matrix of panels, with the last panel missing. The task
is to choose a panel from eight candidates to fill in the missing panel so that three rows form consistent analogies (see Fig. \ref{fig:global} for an example).
To succeed in this task, the model needs to be capable of extracting compositions of three distinct components: (1) the objects in every panel, (2) the visual attributes (such as color, size, and shape) of these objects, and (3) the relationships among attributes, such as ``progression'', ``union'', and ``constant''.


In order to discover rich compositional structures of the data, we propose to build compositions of neural networks. In particular, we compose three types of networks: the object  networks $\{\mathcal{N}^{o}_{i}\}$, the attribute  networks $\{\mathcal{N}^{a}_i\}$, and the relationship networks $\{\mathcal{N}^{r}_i\}$. 
Each composition $\mathcal{N}^{r}_{k}\circ \mathcal{N}^{a}_{j} \circ \mathcal{N}^{o}_{i}$ computes if the $k^{th}$ relationship holds among the $j^{th}$ attributes of the $i^{th}$ objects of the input. We computes all possible compositions of those three neural networks $\{\mathcal{N}^{r}_{k}\circ \mathcal{N}^{a}_{j} \circ \mathcal{N}^{o}_{i}\}_{i,j,k}$ to predict the label. By explicitly computing all possible compositions of three types of neural networks, we force neural networks of each type to be compatible with each other. For example, a relationship neural network that extracts ``progression'' needs to be compatible with all attribute neural networks. Namely, it needs to recognize ``progression'' for any input feature, regardless of whether the input feature represents ``color'', ``size'' or ``shape''. Hence, it is encouraged to learn the exact notion of ``progression'' (a process of gradual advancement), 
instead of an attribute-dependent concept (e.g., ``color getting darker''). The same reasoning applies to all of the networks of three types.
Therefore, the network of each type learns precisely its intended functionality, and the underlying compositional structures are discovered. 

\begin{figure}[t]
  \begin{center}
  \includegraphics[width=0.85\linewidth]{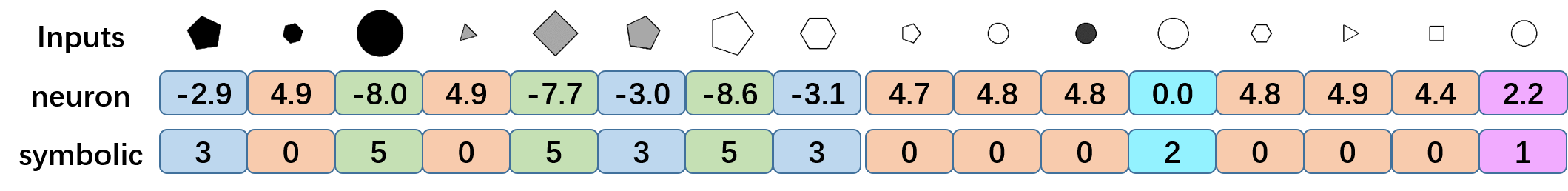}
  \end{center}
  \vspace{-1em}
  \caption{\small SCL learned a ``size neuron'' that is highly predictive of the input object size. From top to bottom: input images from RAVEN, the neuron's activations of each image, the labels of symbolic representation of ``size'' of each image. The neuron learned to represent ``size'' by an approximate linear transformation: $y\approx-\frac{5}{2}x+5$.
  %
  }
  \label{fig:neuron}
  \vspace{-1em}
\end{figure}

We term our proposed model Scattering Compositional Learner (SCL). 
Our SCL achieved the state-of-the-art in two datasets on the RPM task: a 48.7\% relative improvement on Balanced-RAVEN~\cite{balanced_raven} and 26.4\% on PGM~\cite{Santoro2018MeasuringAR} over the previous state-of-the-art. We further show SCL trained with standard end-to-end backpropagation discovers interpretable factorized structures on the Balanced-RAVEN dataset. We found SCL learns to extract symbolic features such as the shape, color, and size for each object. 
We highlight an example of the learned representation in Figure \ref{fig:neuron}, which shows the responses of one of the learned neurons to a set of inputs. This neuron learned a linear transformation of the scalar representation of the symbolic attribute ``size'', captured by the equation $y\approx -\frac{5}{2}x+5$.
As a consequence of learning compositional structures, we further demonstrate that our model is also capable of generalizing to novel test distributions, whereas the previous state-of-the-art model suffered severely from the distribution shift. 

\section{Related Works}{
\label{sec:related}
There has been an old debate over the limits of connectionism for compositional generalization. In a set of influential and
controversial papers, Jerry Fodor and other researchers criticized neural networks for failing to model the intricacy of mind because they cannot capture systematic compositionality~\cite{fodor_connectionism_1988,Marcus1998RethinkingEC, Marcus2001TheAM, Fodor2002TheCP}. These issues have received less attention since deep learning revolutionized various domains.
However, there have been a few papers recently bringing up these issues again, after observing non-robustness in the systematic generalization of neural networks~\cite{Lake2018GeneralizationWS, Loula2018RearrangingTF, Bahdanau2019SystematicGW, Bahdanau2019CLOSUREAS, Keysers2020MeasuringCG}. Our work is an attempt to address the issues of systematic generalization with an architectural bias that respects compositions. We advocate for the same principle as those methods in the past that add modularity for better generalization, such as Neural Module Networks(NMN)~\cite{Andreas2015NeuralMN}, NTPT~\cite{Gaunt2017DifferentiablePW}, 
MAC~\cite{hudson2018compositional}, Neural Symbolic Concept Learner~\cite{mao2018the}, CRL~\cite{Chang2019AutomaticallyCR}. Our work is differentiated from the past work, in that our architecture is specifically designed for learning compositions of a sequence of neural modules, whereas the past work relied on symbolic information (e.g., language in NMN, NSCL, MAC; source code in NTPT), or non-differentiable search (e.g., NSCL, CRL) to compose neural modules in task-specific layout. A more detailed comparison of these approaches is provided in Appendix~\ref{appendix:related}.



SCL also shares some similarities with the group convolution architecture~\cite{alexnet} and its followup work such as ResNeXt~\cite{Xie2016}. Similarly, they perform the ``split'' operation, dividing the computation into multiple groups. However, the transformations in these architectures do not share parameters among groups, whereas ``share'' is a key component in our architecture (see Scattering Transformation in Section~\ref{sec:method}). We present ablation studies to show that our model without parameter sharing could not learn at all (in Appendix \ref{appendix:ablation}), demonstrating the importance of this distinction.


Raven's Progressive Matrices~\cite{raven1936performances} (RPM) are studied extensively in cognitive science~\cite{Carpenter1990WhatOI}. 
It is considered as a diagnostic of human's abstract and structural reasoning ability~\cite{Carpenter1990WhatOI}, and characterizes a central feature of high-level intelligence, namely fluid intelligence~\cite{Jaeggi6829}. 
In the machine learning community, following on the work by \citet{Wang2015AutomaticGO}, \citet{Santoro2018MeasuringAR} introduced the Procedurally Generating Matrices (PGM) dataset of RPM-like problems. They proposed an advanced version of Relational Network~\cite{Santoro2017ASN}, Wild Relational Network (WReN), and studied its generalization on this dataset. Followup work~\cite{Hill2019LearningTM} showed the importance of candidate sets in learning analogies. 
\citet{zhang2019raven} extended PGM and created the dataset RAVEN, with a larger number of relationships in each problem, a task that is considered to be harder than PGM in symbolic reasoning. Several methods have been proposed since then to solve these tasks, including~\cite{LEN_NIPS2019},~\cite{copinet} and~\cite{DBLP:conf/iclr/WangJL20}. However, recently, \citet{balanced_raven} pointed out a short-cut solution in RAVEN which exploited artifacts in the candidate generation procedure. They then created an unbiased version called Balanced-RAVEN which eliminated these artifacts; this is the dataset we have conducted most of our experiments on. 

Lastly, copycat~\cite{copycat} is a cognitive architecture designed to discover insightful analogies, a classical AI approach for analogical reasoning. Its followup Phaeaco~\cite{Phaeaco} was able to demonstrate limited successes in a problem similar to RPM -- the Bongard problems~\cite{Bongard}. These systems use hand-crafted features and patterns detectors, hence they require heavy engineering efforts to scale to challenging analogical reasoning tasks.


}
\section{Methods}
\label{sec:method}

\begin{figure*}[t]
    \centering
    \includegraphics[width=0.98\linewidth]{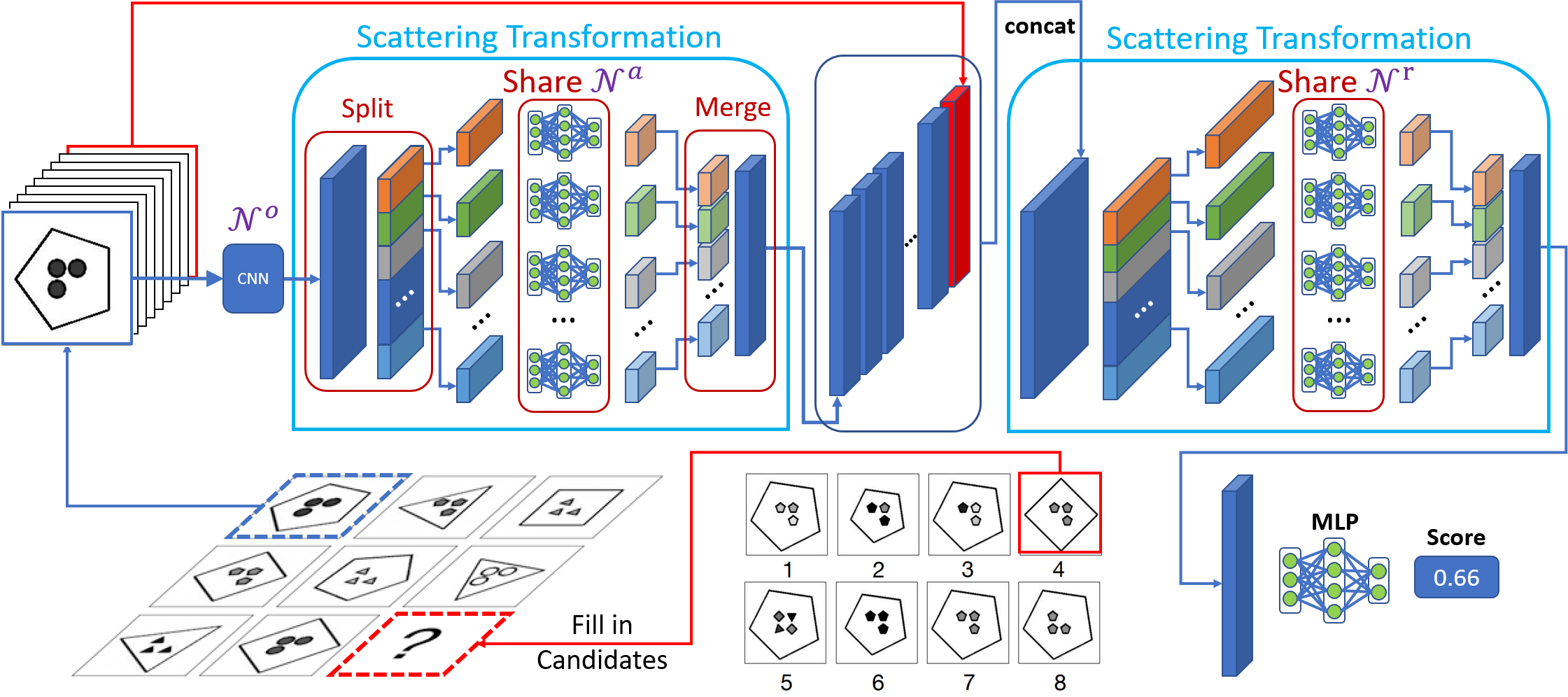}
    \caption{An illustration of the proposed architecture SCL with the scattering transformation.}
    \label{fig:global}
    \vspace{-1em}
\end{figure*}

The Raven's Progressive Matrices (RPM) task~\cite{raven1936performances,John2003} contains rich compositional structures.
To illustrate, in RAVEN~\cite{zhang2019raven}, one is given a $3\times 3$ matrix of panels, with the last panel missing. Each panel is an image that contains a number of objects with visually discernible attributes (such as color, size, and shape). The task is to choose a panel from eight candidates to fill in the missing panel so that three rows form consistent analogies (see Figure \ref{fig:global} for an example).
To succeed in this task, the model needs to be capable of extracting a composition of three distinct components. It first needs to extract objects in every panel. It then needs to extract the visual attributes of these objects. Lastly, it needs to recognize the relationships among attributes, such as ``progression'', ``union'', and ``constant''. In this section, we describe an architecture that is designed to discover the underlying compositional structure of the data.

\subsection{The Scattering Compositional Learner}

In order to discover the underlying compositional structure of the data, we employ three types of neural networks:  object networks $\{\mathcal{N}^{o}_{i}\}$, attribute networks $\{\mathcal{N}^{a}_i\}$, and relationship networks $\{\mathcal{N}^{r}_i\}$. The object networks are convolutional neural networks that extract object representations of from panels; the attribute networks extract attribute features (such as color, size, and shape) from a object representation; the relationship network decides whether certain relationship holds among attributes. 
We explicitly compute the compositions of all these neural networks, $\{\mathcal{N}^{r}_{k}\circ \mathcal{N}^{a}_{j} \circ \mathcal{N}^{o}_{i}\}_{ijk}$ (e.g., whether the $k^{th}$ relationship holds among the $j^{th}$ attributes of the $i^{th}$ objects of the input images), and aggregate the results using an output network for making the prediction. 

By explicitly computing all possible compositions of three types of neural networks, we force the networks of each type to be compatible with each other. For example, a relationship network that extracts ``progression'' needs to be compatible with all attribute networks. Namely, it needs to recognize ``progression'' for any input feature, regardless of whether the input feature represents ``color'', ``size'' or ``shape''. Hence, it is encouraged to learn the general notion of ``progression'' (a process of gradual advancement), 
instead of an attribute-dependent concept (e.g., ``color getting darker''). The same reasoning applies to all neural networks of three types.
As a result, each type of neural network learns precisely its intended functionality, and the underlying compositional structure can be discovered. Furthermore, since the neural networks of different types are learned to be compatible with each other, unseen analogies can also be grasped by novel compositions of these neural networks without further training, as we show in Section~\ref{sec:ood}.

\vspace{-0.5em}
\paragraph{Multi-Head}
Instantiating all neural networks of three types creates nontrivial engineering burdens. Instead, we represent all neural networks of the same type by one multi-head architecture neural network. For example, we use the multi-head attribute neural network $\mathcal{N}^{a}$ with $M$ heads, to represent all attribute neural networks $\{\mathcal{N}^{a}_i\}_{i=1}^M$. Namely, we treat the output of the neural network $\mathcal{N}^{a}$ as the concatenation of the outputs of $\{\mathcal{N}_i^{a}\}_{i=1}^M$. As a result, 
we only need to instantiate $3$ multi-head neural networks in total, instead of $3M$ distinct neural networks.

\vspace{-0.5em}
\paragraph{Scattering transformation} 
With the multi-ahead architecture, we propose an operation called the \emph{scattering transformation}\footnote{The name is inspired by the scattering transform~\cite{scattering}.}  to compute the all possible compositions of neural network components in one single forward pass, largely reducing the computational overhead. 
We define the scattering transformation of an output vector (from $\mathcal{N}^{o}$) by the attribute neural network $\mathcal{N}^{a}$ with the following three steps: (1) \emph{Split} the output vector from $\mathcal{N}^{o}$ into $M$ groups. (2) \emph{Share} the neural transformation $\mathcal{N}^{a}$ for each group. (3) \emph{Merge} all previous results into a single vector. An illustration is shown in Figure \ref{fig:global}. The first step corresponds to splitting $\mathcal{N}^{o}$ into multiple networks $\mathcal{N}^{o}_i$. The second step computes all compositions of $\mathcal{N}^{o}_i$ and $\mathcal{N}^{a}$, where $\mathcal{N}^{a}$ is implicitly a multi-head architecture that computes the concatenation of $\mathcal{N}^{a}_j$. Therefore, the resulting vector after the third step contains outputs from all compositions $\{\mathcal{N}^{a}_j\circ\mathcal{N}^{o}_i\}$. In general, the scattering transformation can be applied repeatedly to compose a sequence of neural network components, as in the case of RAVEN. 


\vspace{-0.5em}
\paragraph{Full Architecture} We describe the full architecture as follows. We first make eight copies of the $3\times 3$ matrix with the missing panel and fill in a different candidate in each of them. The job of SCL is to choose the correctly filled matrix out of eight copies. Every panel of the eight matrices is fed into $\mathcal{N}^{o}$. We perform the first scattering with $\mathcal{N}^{o}$ and $\mathcal{N}^{a}$ for every image, obtaining every attribute of every object. We then perform a second scattering for the output of the first scattering with $\mathcal{N}^{r}$. In the second scattering, since the relationship is defined over the matrix, we feed a concatenation of the attributes of all panels in the matrix into $\mathcal{N}^{r}$ to extract the relationship. Lastly, we aggregate the results by an output MLP that produces a score for each matrix. The predicted probability is computed by taking a softmax over the 8 scores. 
The full diagram is shown in Figure \ref{fig:global}.

\section{Experiments}
\vspace{-0.25em}
We design experiments to answer the following questions: \begin{enumerate*}[label=\textbf{\arabic*)}]
\item \textbf{I.I.D. Generalization:} How does SCL perform on analogical reasoning tasks compared to existing methods? \item \textbf{Learning compositional Structure:} Can SCL discover the underlying compositional structure of the task? How strongly correlated is the learning of compositional structures and the test accuracy?\item \textbf{Compositional generalization:} Can SCL generalize to novel analogies not seen at training time? 
The implementation of the architecture and the experiments is available at \url{https://github.com/dhh1995/SCL}.

\end{enumerate*}

\vspace{-0.25em}
\subsection{Dataset}
\vspace{-0.25em}
\paragraph{PGM} PGM~\cite{Santoro2018MeasuringAR} is a visual analogical reasoning task. The task mimics how Raven Progressive Matrices were designed as a test for Intelligence Quotient (IQ) for humans~\cite{raven1936performances}. 
PGM consists of 8 different subdatasets, each containing 1,222,000 questions with 119,552,000 images. Due to the size of the dataset, we conduct experiments on only one of the sub-datasets, ``neutral''. Each image may contain a variable number of objects of type ``shape'' or ``line''. Five relationships (progression, XOR, OR, AND, and union) are defined over 5 attributes (size, type, color, position, and number) of the objects. There are on average 1-2 relationships defined in each problem.

\vspace{-0.5em}
\paragraph{RAVEN/Balanced-RAVEN} RAVEN~\cite{zhang2019raven} is an extension to the PGM dataset. 
While RAVEN shares the same set of attributes as PGM, it introduces a different set of relationships over these attributes: progression, constant, union and arithmetic. The dataset is designed to be a more difficult task than PGM, as each problem contains 6.3 relationships on average, compared to 1-2 in PGM.
The dataset consists of 7 distinct configurations: \texttt{Center}, \texttt{2x2Grid}, \texttt{3x3Grid}, \texttt{L-R}, \texttt{U-D}, \texttt{O-IC}, \texttt{O-IG}. Each contains 10,000 problems. An illustration of each configuration is shown in Appendix~\ref{appendix:dataset} along with a detailed explanation. 
We evaluated our method on this dataset in addition to PGM because RAVEN also provides labeled symbolic representation of the attributes. This helps us verify the learning of compositional structures in Section \ref{sec:symbolic_concepts}. The symbolic representation is an integer-valued scalar for each feature, ranging from 0 to 9. 

Recently, ~\citet{balanced_raven} observed that there are defects in the design of candidate set generation in RAVEN dataset. In particular, they discovered a short-cut solution that predicts the label perfectly solely based on the candidate set without any information of the context images. Hence, they introduced a revised dataset called Balanced-RAVEN, which generates the candidate set in a different way to eliminate the short-cut solution. 
Since the results on RAVEN can be untrustworthy, in the following experiments, we mainly compare results on Balanced-RAVEN. Results on the original RAVEN are also included for completeness in Appendix~\ref{appendix:raven}. 


\begin{table*}[t]
  \caption{Results on Balanced-RAVEN dataset (joint training).}
  \label{tab:joint_b_raven}
  \centering
    \begin{adjustbox}{width=0.7\linewidth}
  \begin{tabular}{lllllllll}
    \toprule
    &\multicolumn{7}{c}{Test Accuracy($\%$)}\\
    \cmidrule(r){2-9}
    Model     & Avergae &Center    &2Grid    &3Grid    &L-R    &U-D    &O-IC    &O-IG \\
    \midrule
    {\bfseries \textit{LSTM}}~\cite{balanced_raven}      &18.9     &26.2 &16.7 &15.1 &14.6 &16.5 &21.9 &21.1 \\
    {\bfseries \textit{WReN}}~\cite{balanced_raven}     &23.8 &29.4 &26.8 &23.5 &21.9 &21.4 &22.5 &21.5 \\
    {\bfseries \textit{Resnet}}~\cite{balanced_raven}  &40.3 &44.7 &29.3 &27.9 &51.2 &47.4 &46.2 &35.8 \\
    {\bfseries \textit{Wild ResNet}}~\cite{balanced_raven}      &44.3 &50.9 &33.1 &30.8 &53.1 &52.6 &50.9 &38.7 \\
    {\bfseries \textit{LEN}}~\cite{LEN_NIPS2019} & 39.0 &45.5   & 27.9 & 26.6   &44.2   &43.6 &50.5   & 34.9\\
    {\bfseries \textit{CoPINet}}~\cite{copinet} &46.3 &54.4 & 33.4 & 30.1 & 56.8 & 55.6 & 54.3 & 39.0\\
    {\bfseries \textit{HriNet}}~\cite{balanced_raven} &63.9 &80.1 &53.3 &46.0 &72.8 &74.5 &71.0 &49.6\\
    {\bfseries \textit{SCL~(ours)}}         &\textbf{95.0}    &\textbf{99.0}    &\textbf{96.2}        &\textbf{89.5}    &\textbf{97.9}    &\textbf{97.1}    &\textbf{97.6}    &\textbf{87.7}    \\
    \midrule 
    {\bfseries \textit{Human}}\footnotemark[2] 
    ~(\cite{zhang2019raven})&84.4    &95.5    &81.8    &79.6    &86.4    &81.8    &86.4    &81.8    \\
    \bottomrule
  \end{tabular}
  \end{adjustbox}
  \vspace{-1em}
\end{table*}

\footnotetext[2]{The human experiments were conducted on the original RAVEN dataset, obtained from~\citet{zhang2019raven}.}

\vspace{-0.25em}
\subsection{Experimental Protocol}
\vspace{-0.25em}
\label{sec:train}
For Balanced-RAVEN, we used 70,000 examples in total, and used a training, validation, and test split of $60\%$, $20\%$ and $20\%$ following~\citet{zhang2019raven}.
We ran every model for 300 epochs, with 5 random seeds. 
We took the best-validated model out of 5 runs and evaluated on the test set. 
We used the same architectural hyperparameters for our model in all the tasks, as detailed in Appendix \ref{appendix:model_archi}. In training our model, we used the Adam optimizer~\cite{kingma2014adam} with a learning rate of 0.0007 for PGM and 0.005 for Balanced-RAVEN/RAVEN, while otherwise keeping the default Adam hyperparameters. We also applied a weight decay of 0.01. The hyperparameters were found after a search over the learning rate from $\{0.007, 0.005, 0.001, 0.0007, 0.0005\}$ and weight decay from $\{0.0, 0.001, 0.01\}$. In all of our experiments, we used one Nvidia P100 GPU with 12 GB RAM with 8 CPU cores.


\vspace{-0.25em}
\subsection{I.I.D. Generalization Tasks}
\vspace{-0.25em}
\label{sec:iid}
In this section, we focus on i.i.d.~generalization, where the training data and the test data come from the same distribution. 
As we focus on architectural comparisons in the following evaluations, we consider the following baselines: LSTM \cite{Hochreiter1997LongSM}, WReN~\cite{Santoro2018MeasuringAR}, CNN+MLP, ResNet-18~\cite{He2015},
MXGNet~\cite{DBLP:conf/iclr/WangJL20},
LEN~\cite{LEN_NIPS2019}, T-LEN~\cite{LEN_NIPS2019}, CoPINet~\cite{copinet}, HriNet~\cite{balanced_raven} \footnote[3]{We took the implementation of LSTM, CNN+MLP, and ResNet-18, WReN from \url{https://github.com/Fen9/WReN}, implementation of LEN from \url{https://github.com/zkcys001/distracting_feature} and implementation of CoPINet from \url{https://github.com/WellyZhang/CoPINet}.}. We do not consider baselines that make use of auxiliary labels, or techniques for optimizing the training data. We were not able to run MXGNet on Balanced-RAVEN due to the lack of an open-source implementation. 

\vspace{-0.25em}
\subsubsection{PGM}
\vspace{-0.25em}
\begin{wraptable}{r}{0.35\textwidth}
  \vspace{-3em}
  \caption{Results on PGM dataset.}
  \label{tab:pgm}
  \centering
  \begin{adjustbox}{width=0.85\linewidth}
  \begin{tabular}{ll}
    \toprule
    Model& Acc(\%)\\
    \midrule
    {\bfseries \textit{LSTM}}~(\cite{Santoro2018MeasuringAR})      &33.0  \\
    {\bfseries \textit{CNN+MLP}}~(\cite{Santoro2018MeasuringAR})      &35.8  \\
    {\bfseries \textit{Resnet-50}}~(\cite{Santoro2018MeasuringAR})  &42.0  \\
    {\bfseries \textit{W-ResNet-50}}~(\cite{Santoro2018MeasuringAR}) &48.0 \\
    {\bfseries \textit{WReN}}~(\cite{Santoro2018MeasuringAR})  &62.8 \\
    {\bfseries \textit{MXGNet}}~(\cite{DBLP:conf/iclr/WangJL20})  &66.7 \\
    {\bfseries \textit{CoPINet}}~(\cite{copinet})       &56.4  \\
    {\bfseries \textit{LEN}}~(\cite{LEN_NIPS2019})       &68.1  \\
    {\bfseries \textit{T-LEN}}~(\cite{LEN_NIPS2019})       &70.3  \\
    \midrule
    {\bfseries \textit{SCL~(ours)}}         &\textbf{88.9}   \\
    \bottomrule
  \end{tabular}
  \end{adjustbox}
  \vspace{-2em}
\end{wraptable}
We followed the training setting and experimental protocol described in Section \ref{sec:train}, and the results are shown in Table \ref{tab:pgm}. One can see that our SCL achieved a substantial improvement over the baselines. The SCL achieved a relative improvement of $26.4\%$ over the previous state-of-the-art method T-LEN.

\vspace{-0.25em}
\subsubsection{Balanced-RAVEN/RAVEN}
\vspace{-0.25em}
\label{sec:raven_main}
Due to the aforementioned shortcut solution in RAVEN, we evaluated all methods on the unbiased Balanced-RAVEN dataset. \citet{zhang2019raven} introduced the \textit{joint} training setting, where the model is trained and evaluated on all configurations jointly. The results are shown in Table \ref{tab:joint_b_raven}. 
One can see that our proposed method SCL achieved a substantial improvement over the baselines. On average, SCL achieved $31.1$\% absolute improvement over the previous state-of-the-art HriNet, from $63.9\%$ to $95.0\%$ in test accuracy, which is a $48.7\%$ relative improvement. 
In addition, we also introduce \textit{single} training setting, where the model is trained and evaluated on each configuration separately. Due to the lack of space, the results are shown in the Appendix, Table \ref{tab:single_b_raven}. Our model achieves perfect $100\%$ test accuracy on 4 out of 7 configurations, significantly outperforming the baselines, which obtain test accuracy around $50\%$. 

\vspace{-0.25em}
\subsubsection{Which Methods Exploit the Shortcut Solution?}
\label{sec:short_cut}
\vspace{-0.25em}
\begin{wraptable}{r}{0.35\textwidth}
  \vspace{-1.3em}
  \caption{Sanity check on RAVEN when context images are masked out.}
  \label{tab:mask}
  \centering
  \begin{adjustbox}{width=0.85\linewidth}
  \begin{tabular}{lll}
    \toprule
    Model \textbackslash Context& Yes & No\\
    \midrule
    {\bfseries \textit{CoPINet}}~(\cite{copinet})      & 91.4 & 95.0 \\
    {\bfseries \textit{SCL}~(ours)}~    &     91.6 & 12.2 \\ 
    \bottomrule
  \end{tabular}
  \end{adjustbox}
  \vspace{-0.3em}
\end{wraptable}
As pointed out by ~\citet{balanced_raven}, there exists a shortcut solution to the original  RAVEN dataset. To investigate which methods exploit the shortcut, we first compared the results on RAVEN to Balanced-RAVEN. The results on the original RAVEN dataset are shown in Appendix \ref{appendix:raven}. We observed that most of the existing baselines' performances degraded significantly. 
In particular, the previous state-of-the-art method CoPINet~\cite{copinet}, achieving a test accuracy of 91.4$\%$ on RAVEN, was only able to obtain a test accuracy of 46.3$\%$ on Balanced-RAVEN. This indicates that the existing baselines very likely exploited the short-cut solution. In contrast, the performance of SCL was robust to the change in candidate generation (we will also show how SCL solved the task exactly in the following section). We further verified this by masking out all 8 panels that provide contextual cues to the correct answer on RAVEN. We found CoPINet achieved an even higher test accuracy of 95.0$\%$, suggesting that this model exploited the short-cut solution~\cite{balanced_raven} to obtain good performances on this task, whereas our model fell to 12.2$\%$ (corresponding to chance accuracy), as shown in Table~\ref{tab:mask}.

\subsection{Does SCL Discover Compositional Structure?}
\label{sec:symbolic_concepts}
Encouraged by the results shown in the previous section, we analyzed SCL to understand the source of the improvements over the baselines. 
We examined whether SCL learned the underlying compositional structures of the dataset. 
We conducted our analysis on Balanced-RAVEN instead of PGM, because RAVEN contains the labels of symbolic attributes of every object, which is crucial to verify whether it discovered compositional structures. 
In the following sections, we first investigate the learning of objects and their attributes by examining the outputs of neural networks $\mathcal{N}^{o}$ and $\mathcal{N}^{a}$. We then investigate whether SCL learned to identify the symbolic relationship by examining $\mathcal{N}^{r}$.


\subsubsection{Learning Objects and Attributes}
Since the dataset provides the ground truth symbolic attributes of every object for each scene, we can examine whether the composition of neural networks $\mathcal{N}^{o}$ and $\mathcal{N}^{a}$ learned to extract attributes (i.e., ``color'', ``type'', and ``size'') of each object of the scene. Formally, we denote the ground truth $i^{th}$ attribute of the $j^{th}$ object on an image $X$ by $f^{a}_i\circ f^{o}_{j}(X)\in\mathbb{R}$. We let $\mathcal{N}^{o}_j$ represent the extraction of $j^{th}$ \textit{object}, and $\mathcal{N}^{a}_i$ represent the extraction of $i^{th}$ \textit{attribute}. We examine if the ground truth symbolic attributes match any of the compositions of neural modules $\{\mathcal{N}^{a}_i\circ\mathcal{N}^{o}_j\}_{i,j}$ up to permutation of the indices, followed by a linear transformation. Namely, we say that the model learned to extract the $i^{th}$ attribute of $j^{th}$ object, $f^{a}_i\circ f^{o}_{j}$, if there exists an linear transformation that maps a neural composition $\mathcal{N}^{a}_q \circ \mathcal{N}^{o}_p$ to the ground truth for every image $X$ approximately,
\begin{equation*}
     f^{a}_i\circ f^{o}_{j}(X)\approx \mathrm{Linear}( \mathcal{N}^{a}_q\circ\mathcal{N}^{o}_p(X); W,b), \forall X.
\end{equation*}

Therefore, to verify if the model learned the composition $(i,j)$, we took the learned model and searched for the linear transformation and the pair by optimizing the following $(i, j)$-composition loss:
%
\begin{equation*}
    \mathcal{L}_{comp}(i,j) =  \min_{\substack{W, b\\ p,q\in[1,\dots,M]^2}} \frac{1}{|\mathcal{D}|}\sum_{X\in\mathcal{D}} ||\mathrm{Linear}( \mathcal{N}^{a}_q\circ\mathcal{N}^{o}_p(X); W,b) -f^{a}_i\circ f^{o}_{j}(X)||^2.
\end{equation*}
\begin{figure*}[t]
  \centering
  \begin{subfigure}{0.45\textwidth}
  \centering
  \vspace{0.3em}
  \includegraphics[width=0.98\textwidth]{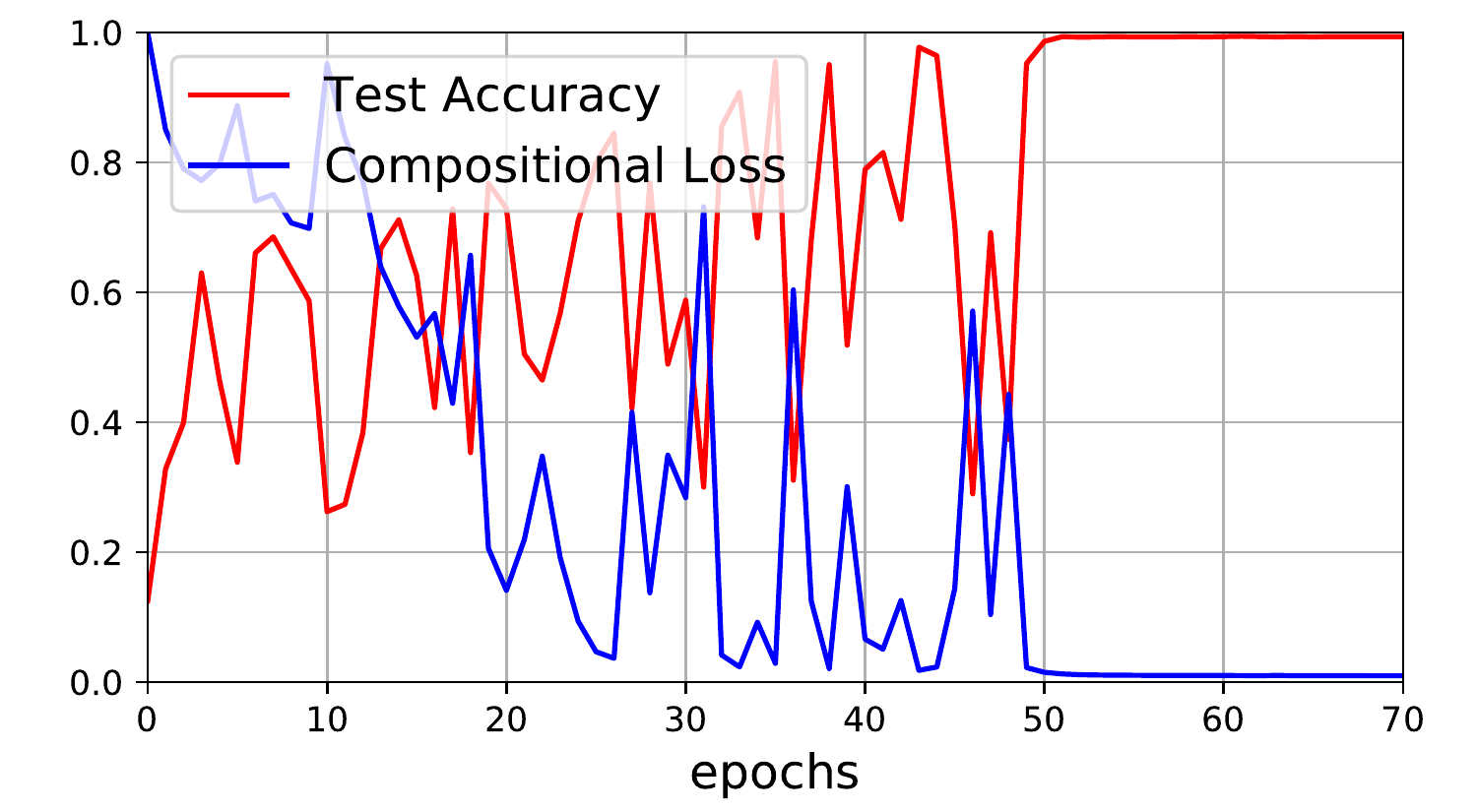}
  \vspace{0.3em}
  \caption{We show test accuracy and $\mathcal{L}_{comp}$ in task \texttt{L-R}. The test accuracy is a mirror image of the $\mathcal{L}_{comp}$, suggesting learning compositional structures is strongly correlated with generalization.}
  \label{fig:concept}
  \end{subfigure}\quad\quad
  \begin{subfigure}{0.45\textwidth}
  \centering
  \includegraphics[scale=0.16]{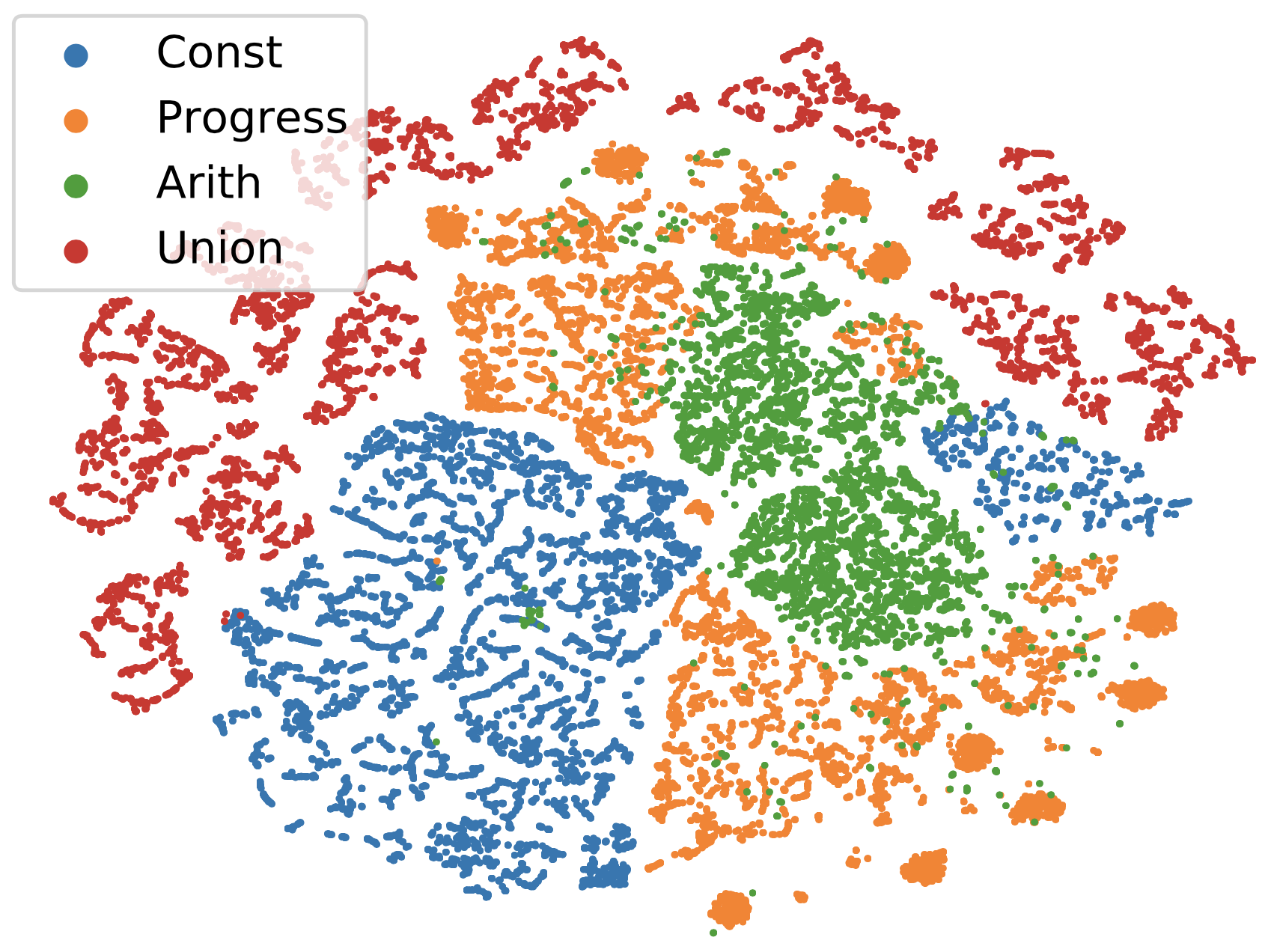}
  \caption{t-SNE plot of the output vectors of the relationship neural network $\mathcal{N}^{r}$ in task \texttt{L-R}. Each cluster has clear boundary, indicating the neural network was able to identify every relationships clearly.}
  \label{fig:tsne}
  \end{subfigure}
  \caption{Visualization of compositional structure learning.}
  \vspace{-1em}
\end{figure*}

\begin{table}[htb]
  \vspace{-1.0em}
  \caption{Test Compositional Structure Loss and the symbolic representation prediction accuracy for the model trained on joint tasks.}
  \vspace{0.3em}
  \label{tab:concept}
  \centering
  \begin{adjustbox}{width=0.6\linewidth}
  \begin{tabular}{llllllll}
    \toprule
    \cmidrule(r){2-8}
          &Center    &2Grid    &3Grid    &L-R    &U-D    &O-IC    &O-IG \\
    \midrule
    {\bfseries \textit{Loss}}
    &0.02    &0.27   &0.35    &0.04    &0.05    &0.04   &0.26 \\
     {\bfseries \textit{Acc}}         
     &99.4    &85.8    &79.8    &98.4    &96.6    &97.9   &80.5 \\
    \bottomrule
  \end{tabular}
  \end{adjustbox}
\end{table}

We took the model we trained in Section \ref{sec:iid}, and recorded the average of compositional learning losses for attributes ``color'', ``shape'', and ``type'' of every object in the test data. In addition to the loss, we also measured the accuracy of predicting the symbolic representation by the linear transformation followed by rounding.
The loss as well as the accuracy for each configuration 
is reported in Table \ref{tab:concept}. 
We can see that in configurations \texttt{Center}, \texttt{L-R}, \texttt{U-D}, and \texttt{O-IC}, the model learned the attributes almost perfectly, with close to full predictability. 
It learned slightly worse in the other three configurations, echoing the degradation of test accuracy for these tasks. 

We also tracked the compositional loss along training to show the evolution of compositional structure learning. We used the single training setting of \texttt{L-R} for illustration. 
We plotted the average loss of $\mathcal{L}_{comp}$ over all six compositions (3 attributes $\times$ 2 objects) along with training, as well as the test accuracy in Figure \ref{fig:concept}. 
We observed that the test accuracy and $\mathcal{L}_{comp}$ evolved exactly in the opposite trends of each other at every epoch. 
This strong correlation supports the hypothesis that the success of generalization was largely due to learning compositional structures.

\subsubsection{Learning Relationships}
Lastly, we investigated if the model learned to extract meaningful relationships. Given the ground truth relationship defined on each attribute, we performed a t-SNE~\cite{vanDerMaaten2008} for the output vectors of $\mathcal{N}^{r}$ over 1024 examples, each containing $4$ relationships. An illustration with configuration \texttt{L-R} is shown in Figure \ref{fig:tsne}. We observed that each cluster is formed by a single relationship, with very little overlap, indicating the model was able to identify each relationship concept. 
We also visualized the relationship learning in all configurations via t-SNE in Appendix \ref{appendix:tsne}, where we observed clusters were well-separated in 4 out of 7 tasks. 

\subsection{Generalization to Unseen Analogies}
\label{sec:ood}

If the model understands attribute-relationship pair (``color'', ``constant''), and (``size'', ``progression''), can the model generalize to (``color'', ``progression'')? We hypothesize that if the model uncovers the compositional structure of attributes and relationships, new combinations of those known concepts can be grasped without further training.
For a given relationship-attribute pair, we created a dataset where the pair appear in every test examples but not in the training/validation set. 
We examined the joint training setting, and took 9 pairs that exist in all configurations and created a new dataset for each of them. 

We compared SCL to CoPINet, the previous state-of-the-art on RAVEN.
The test accuracy and the generalization gaps (i.e., the difference between the test accuracy and the validation accuracy) are shown in Table \ref{tab:joint_ood}. SCL achieved an average test accuracy of $90.0\%$ over 9 novel attribute-relationship pairs, whereas the CoPINet only achieved an average of $34.7\%$. We also observe SCL had an average degradation of $2\%$ from validation to test accuracy over nine pairs, whereas CoPINet heavily suffered from distribution shift, with a degradation of $14.6\%$ on average. We also examined the single training setting with the configuration \texttt{L-R}, where results are reported in the Appendix~\ref{appendix:novel_pairs_lr}. 

We additionally created another task for testing out-of-distribution generalization, where the model is asked to generalize to a different number of attribute-relationship pairs (see Appendix~\ref{appendix:num_rels} for details). Similarly to our previous findings, SCL models were able to generalize but the baseline models failed badly. 





\begin{table}[t]
\centering 
\caption{\small Generalization to unseen attribute-relationship pairs for joint training data.}
  \label{tab:joint_ood}
  \begin{adjustbox}{width=\linewidth}
  \begin{tabular}{llllllll}
    \toprule
    &\multicolumn{6}{c}{Test Accuracy (Difference to Validation Accuracy)}\\
    \cmidrule(r){2-7}
     & \multicolumn{2}{c}{Type} & \multicolumn{2}{c}{Size} & \multicolumn{2}{c}{Color} \\
     &{\bfseries\textit{CoPINet}}~\cite{copinet}
     & {\bfseries\textit{SCL}}~(ours)
     &{\bfseries\textit{CoPINet}}~\cite{copinet}
     &{\bfseries\textit{SCL}}~(ours)
     &{\bfseries\textit{CoPINet}}~\cite{copinet}
     &{\bfseries\textit{SCL}}~(ours) \\
    \midrule
    {Constant}         & 21.7 (-25.5) & \textbf{90.2 (-0.2)}  & 39.5 (-10.2) & \textbf{92.1 (-0.3)}  & 37.0 (-6.6) & \textbf{83.8 (-8.0)} \\ 
    {Progression}      & 33.6 (-20.7) & \textbf{90.1 (-2.1)}  & 42.9 (-5.1) & \textbf{90.3 (-0.7)} & 40.8 (-7.0) & \textbf{93.3 (+0.1)} \\ 
    {Union}            & 32.9 (-22.5) & \textbf{85.4 (-6.9)}  & 34.1 (-17.0) & \textbf{92.0 (-1.0)}  & 29.8 (-17.1) & \textbf{92.5 (+0.1)} \\ 

    \bottomrule
  \end{tabular}
  \end{adjustbox}
\end{table}

\vspace{-0.25em}
\section{Conclusion}
\vspace{-0.25em}
In this work, we introduced a neural architecture SCL for discovering the underlying compositional structure of the data. We applied the proposed method to an analogical reasoning task, Raven's Progressive Matrices, which exhibits strong compositional structures. SCL achieved state-of-the-art performance in two datasets, with a 48.7\% relative improvement on Balanced-RAVEN and 26.4\% on PGM. We validated the learned compositional structure by showing that the neural representation matches the symbolic representation of attribute concepts up to a linear transformation, and well-separated relationship clusters in a t-SNE plot. With learned compositional structures, 
SCL significantly outperforms other methods under test-time distribution shifts. Our work provides a promising research direction in designing compositional neural architectures to achieve stronger generalization. 

\bibliographystyle{plainnat}
\bibliography{citation}
\appendix
\appendix
\newpage

\section{Architecture Details}
\label{appendix:model_archi}

The architectural hyperparameter is the following. In our architecture, we use a feedforward residual block, and hence we define it upfront: a feedforward residual block (motivated by~\cite{He2016DeepRL}) denoted by $FR(x) = x + \mathrm{Lin}_2(\mathrm{LayerNorm}(\mathrm{ReLU}(\mathrm{Lin}_1(x)))$, where $\mathrm{LayerNorm}$ denotes Layer Normalization~\cite{ba2016layer}, and $\mathrm{Lin}_1$ and $\mathrm{Lin}_2$ are two linear layers with the same input output size.

To apply the proposed architecture SCL to the RAVEN task, we first make eight copies of the $3\times 3$ matrix with the missing panel and fill in a different candidate in each of them. Given one of the eight matrices, we pass every panel into $\mathcal{N}^{o}$ to extract object representation. $\mathcal{N}^{o}$ consists of a convolutional neural network of four layers, with channel sizes 16, 16, 32, 32, kernel size $3\times 3$ and padding $1$, followed by a linear layer with a Relu activation with an output dimension of 80, followed by a feedforward residual block. We then perform scattering with $\mathcal{N}^{a}$: 1. ``Split'' the output of $\mathcal{N}^{o}$ into 10 heads. 2. ``Share'' the neural network transformation $\mathcal{N}^{a}$ for each head. $\mathcal{N}^{a}$ is a one-layer MLP with hidden size $128$ with output size $8$. 3. ``Merge'': the output of $10$ heads is then concatenated as a vector of $80$ dimension, followed by a feedforward residual block. Note that the last feedforward residual block is applied to the output of all objects' attributes as a whole. The reason is to extract some potentially useful global information (such as ``number'' of objects). Since the transformation is a residual block $x+f(x)$, it also will maintain useful original information
$x$ for independent attributes if it needs to.
We then perform the second scattering transformation with $\mathcal{N}^{r}$: 1.``Split'': we split the $80$ dimensional output vector from the previous scattering transformation into $80$ groups. We concatenate 1 scalar of each panel from the entire matrix to form a vector of 9 dimensions.  2. ``Share'': We take each $9$ dimensional vector as the input to $\mathcal{N}^{r}$, which is a two-layer MLP with hidden sizes 64 and 32, with an output size of $5$. 3.``Merge'': concatenate 80 groups of $5$ neurons into a vector of $400$. Lastly, each $400$ dimensional vector is fed into an MLP of hidden size 128 to obtain a score for each matrix. An illustration of the diagram is found in Figure \ref{fig:global}.

\section{Details on the RAVEN dataset}
\label{appendix:dataset}
\begin{figure*}[h]
    \centering
    \includegraphics[width=0.98\linewidth]{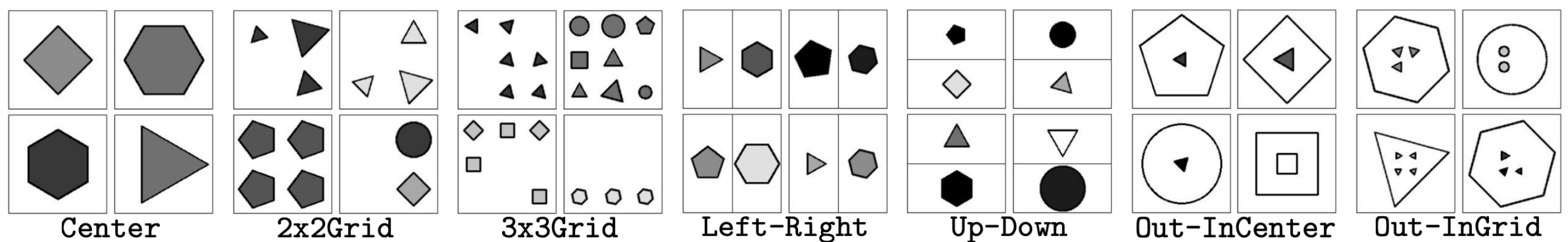}
    \caption{Examples of 7 different figure configurations in the RAVEN dataset (taken from \cite{zhang2019raven}).}
    \label{fig:raven}
    \vspace{-1em}
\end{figure*}

The configuration \texttt{Center} is the simplest one, where there is a single object in the center of each panel. The analogies are constructed by defining one relationship on every object attribute, totaling to three analogies in each problem. In configurations \texttt{L-F}, \texttt{U-D}, \texttt{O-IC}, each image contains two objects of various placements. The analogy is defined on every attribute of each object in the panel, in a total of six analogies. When there are more objects, in configurations such as \texttt{2x2Grid}, \texttt{3x3Grid}, more sophisticated analogies are introduced, such as ``progression" over ``number" and ``arithmetic" over ``position". An illustration of each configuration is show in Figure \ref{fig:raven}.

\section{Further Results on RAVEN/Balanced-RAVEN}
\subsection{Results on Balanced-RAVEN}
We also introduce single-task training settings, where the model was trained and evaluated on each configuration separately. Due to the lack of space, the results are shown in Table \ref{tab:single_b_raven}. SCL can achieve nearly perfect test accuracy on 4 out of 7 configurations.

\begin{table*}[h]
  \caption{Single Task Training results on Balanced-RAVEN dataset.}
  \label{tab:single_b_raven}
  \centering
    \begin{adjustbox}{width=0.7\linewidth}
  \begin{tabular}{lllllllll}
    \toprule
    &\multicolumn{7}{c}{Test Accuracy($\%$)}\\
    \cmidrule(r){2-9}
    Model     & Avergae &Center    &2Grid    &3Grid    &L-R    &U-D    &O-IC    &O-IG \\
    \midrule
    {\bfseries \textit{LSTM}}(~\cite{zhang2019raven})
    &12.5    &12.3    &13.3    &12.8    &12.7    &10.3    &12.9    &13.1 \\
    {\bfseries \textit{WReN}}
    &17.8    &23.3    &18.1    &17.4    &16.5    &15.2    &16.8    &17.3 \\
    {\bfseries \textit{CNN+MLP}}
    &12.9    &12.9    &13.2    &12.7    &11.5    &13.5    &12.9    &13.7 \\
    {\bfseries \textit{Resnet-18}}
    &14.5    &20.8    &12.9    &14.3    &13.2    &13.4    &13.8    &12.9 \\
    {\bfseries \textit{LEN}}
    &28.4    &42.5    &21.1   &19.9    &27.6    &28.1    &32.9    &27.0  \\
    {\bfseries \textit{CoPINet}}
    &38.6    &50.4    &30.9   &28.5    &40.0    &40.8    &42.7    &36.9 \\
    {\bfseries \textit{SCL(ours)}} &\textbf{85.4}  & \textbf{99.8} & \textbf{72.4} & \textbf{64.2} & \textbf{99.5} & \textbf{99.4} & \textbf{98.6} & \textbf{64.2} \\
    \bottomrule
  \end{tabular}
  \end{adjustbox}
  \vspace{-1em}
\end{table*}

\subsection{Sanity check on RAVEN}
\label{appendix:raven}
We show further results on RAVEN for completeness. The results of the joint training setting are shown in Table \ref{tab:joint_raven}. By comparing the results on RAVEN to Balanced-RAVEN, we conclude that most of the existing baselines have the issues discovered by~\cite{balanced_raven}, namely, they exploited the short-cut solution, whereas SCL was robust to the change in candidate generation. Notably, the previous state-of-the-art method CoPINet~\cite{copinet}, achieving a test accuracy of 91.4$\%$ on RAVEN, only was able to obtain a test accuracy of 46.3$\%$ on Balanced-RAVEN, strongly suggesting that this model exploited the cheating solution to obtain good performances on this task.

\begin{table*}[h]
  \caption{Joint Training results on RAVEN dataset.}
  \label{tab:joint_raven}
  \centering
  \begin{adjustbox}{width=0.7\linewidth}
  \begin{tabular}{lllllllll}
    \toprule
    &\multicolumn{7}{c}{Test Accuracy}\\
    \cmidrule(r){2-9}
    Model     & Avergae &Center    &2Grid    &3Grid    &L-R    &U-D    &O-IC    &O-IG \\
    \midrule
    {\bfseries \textit{LSTM}}~(\cite{zhang2019raven})      &13.1    &13.2    &14.1    &13.7&    12.8    &12.5    &12.5    &12.9 \\
    {\bfseries \textit{WReN}}~(\cite{zhang2019raven})      &14.7    &13.1    &28.6    &28.3    &7.5    &6.3    &8.4    &10.6 \\
    {\bfseries \textit{CNN+MLP}}~(\cite{zhang2019raven})  &37.0    &33.6    &30.3    &33.5    &39.4    &41.3    &43.2    &37.5 \\
    {\bfseries \textit{Resnet-18}}~(\cite{zhang2019raven}) &53.4    &52.8    &41.9    &44.2    &58.8    &60.2    &63.2    &53.1 \\
    {\bfseries \textit{LEN}}~(\cite{LEN_NIPS2019})       &72.9    &80.2    &57.5    &62.1    &73.5    &81.2    &84.4    &71.5 \\
    {\bfseries \textit{CoPINet}}~(\cite{copinet})       &91.4 &95.1 &77.5 &78.6 &99.1 &99.7 &98.5 &91.4 \\
    {\bfseries \textit{SCL(ours)}}         &91.6    &98.1    &91.0    &82.5    &96.8    &96.5   &96.0    &80.1    \\
    \midrule
    {\bfseries \textit{Human}}~(\cite{zhang2019raven})         &84.4    &95.5    &81.8    &79.6    &86.4    &81.8    &86.4    &81.8    \\
    \bottomrule
  \end{tabular}
  \end{adjustbox}
  \vspace{-1em}
\end{table*}

\section{Comparisons to Previous Modular Neural Networks}
\label{appendix:related}
There has been an old debate over the limits of connectionism for compositional generalization. In a set of influential and
controversial papers, Jerry Fodor and other researchers criticized that neural networks fail to model the intricacy of mind because they cannot capture systematic compositionality~\cite{fodor_connectionism_1988,Marcus1998RethinkingEC, Marcus2001TheAM, Fodor2002TheCP}. Such voices have died down since deep learning revolutionized various domains. However, there have been a few papers recently bringing up these issues again, after observing non-robustness in the systematic generalization of neural networks~\cite{Lake2018GeneralizationWS, Loula2018RearrangingTF, Bahdanau2019SystematicGW, Bahdanau2019CLOSUREAS, Keysers2020MeasuringCG}. Our work is an attempt to address the issues of systematic generalization with an architectural bias that respects compositions. We advocate for the same principle as those methods in the past that add modularity for better generalization, such as Neural Module Networks(NMN)~\cite{Andreas2015NeuralMN}, NTPT~\cite{Gaunt2017DifferentiablePW}, 
MAC~\cite{hudson2018compositional}, Neural Symbolic Concept Learner~\cite{mao2018the}, CRL~\cite{Chang2019AutomaticallyCR}. We detail what each model does and compare it to our models.

\paragraph{Neural Module Networks} NMN is a neural network that is assembled from what are known as neural modules. Each neural module is a neural network that is specialized in some particular subtask. Hence, by composing different neural modules, NMN can perform various tasks, analogous to how a computer program composed of functions. However, how to connect different modules requires domain knowledge. In the original NMN, the layout was set in an ad-hoc manner for each question by analyzing a dependency parse on the language query. Similar to NMN, each neural network of each type can be seen as a neural module. However, our approach is specifically designed for composing a sequence of neural networks, without any help from symbolic hints. 

\paragraph{Neural Symbolic Concept Learner} NSCL extends the NMN work by training the dependency parser jointly with the neural modules, without any auxiliary labels on parsing. Namely, it is a method of learning how to connect neural modules along with training neural modules themselves. As the parser operates on discrete tokens hence indifferentiable, they used REINFORCE to estimate the gradient.  
Despite learning the layout of NMN in an unsupervised manner, NSCL still makes use of language data to guide the construction of the computational graph. In contrast, our model learned compositional structures without any symbolic hints.

\paragraph{MAC}
MAC is a recurrent cell composed of control, read and write units that maintain a separation between control and memory. 
The control units attend to natural language questions and employ the read and write units to gather information and update memory for performing reasoning. By stacking together multiple recurrent MAC cells, the MAC network performs an explicit multi-step reasoning process. 
However, MAC could suffer from handling the long reasoning chain without language guidance in the case of RAVEN. Instead, our SCL learn to reason about the correct answer without any auxiliary labels or hints, solely based on the compositional inductive bias.

\paragraph{NEURAL TERPRET} NTPT provides a system for constructing differentiable program interpreters. In their work, the benefits of knowledge transfer brought by modularity were extensively studied. Similar to the prior work, it requires additional symbolic information (source code) for composing neural modules. 

\paragraph{Compositional Recursive Learner} CRL builds on the prior work and study how to connect neural modules via reinforcement learning (RL). The method was demonstrated on a few simple tasks, but also admitted issues with scalability to more challenging problems due to large search space with RL.

In summary, our work differentiates from the past work, in that our architecture targets at composing a sequence of neural modules. The past work relies on domain knowledge or ad-hoc design (e.g., NMN), and symbolic information (e.g., language in NSCL, MAC; source code in NTPT), or non-differentiable search(e.g., NSCL, CRL) for composing neural modules in task-specific layout.

\section{Ablation Studies}
\label{appendix:ablation}
To understand the way how our model works, we perform the following ablation studies, each starts with a question. For all the experiments run in this section, we follow the experimental protocol described in Section \ref{sec:train}. The results reported in this section used an average of 5 runs.

\paragraph{How important is sharing?} Compared to ResNeXt, a crucial difference is that we used a shared transformation in the \emph{scattering transformation}. To validate the importance of this design choice, we show the average results on the joint training task for sharing versus non-sharing models in Table \ref{tab:sharing}.

\begin{table}[h]
  \caption{How important is sharing? }
  \label{tab:sharing}
  \centering
  \begin{tabular}{llll}
    \toprule
    &\multicolumn{2}{c}{Test Accuracy}\\
         &Sharing &Non-sharing   \\
    \midrule
        &\textbf{95.0}  &12.3   \\
    \bottomrule
  \end{tabular}
\end{table}

\paragraph{How to choose the number of heads of $\mathcal{N}^{a}$?} We took the single training setting of the configuration \texttt{Center} for illustration. Since there are in total 3 attributes in this task (color, size, type), it may suggest using 3 heads for $\mathcal{N}^{a}$ is a good choice on this task. However, this turned out to be a poor choice. The average test accuracy over 5 runs were shown in Table \ref{tab:choose_h}. We found that as we increase the number of heads, it is easier for the model to learn the ground truth solution. We believe this is because of an ensemble effect, where more number of neurons provide more opportunities for learning the correct attribute concepts.
\begin{table}[h]
  \caption{How to choose the number of heads of $\mathcal{N}^{a}$ for \texttt{Center}? }
  \label{tab:choose_h}
  \centering
  \begin{tabular}{lllll}
    \toprule
    &\multicolumn{3}{c}{Test Accuracy}\\
         &4&8&80   \\
    \midrule
        &57.3&43.2&\textbf{100.0}   \\
    \bottomrule
  \end{tabular}
\end{table}

\paragraph{How important is to use scattering transformation with $\mathcal{N}^{o}$?} It may seem plausible that if the convolutional neural network is strong enough, it can extract the attributes of all objects directly without the first scattering transformation of $\mathcal{N}^{o}$ (i.e., when $\mathcal{N}^{o}$ is a regular single-head CNN). We showed in Table \ref{tab:obj} that compares the number of heads of $\mathcal{N}^{o}$ in the joint training setting. 
We observe that having more than 1 head was essential to our model. 

\begin{table}[h]
  \caption{Test accuracy with various number of heads of $\mathcal{N}^{o}$ in joint training setting.}
  \label{tab:obj}
  \centering
  \begin{tabular}{llllll}
    \toprule
    &\multicolumn{4}{c}{Test Accuracy}\\
         &1 &4&10&16   \\
    \midrule
        &12.4&34.5&\textbf{95.0}  &88.2   \\
    \bottomrule
  \end{tabular}
\end{table}



\section{Further Visualizations of Relationship Learning}
\label{appendix:tsne}
\begin{figure}[ht]
    \centering
    \includegraphics[width=0.3\linewidth]{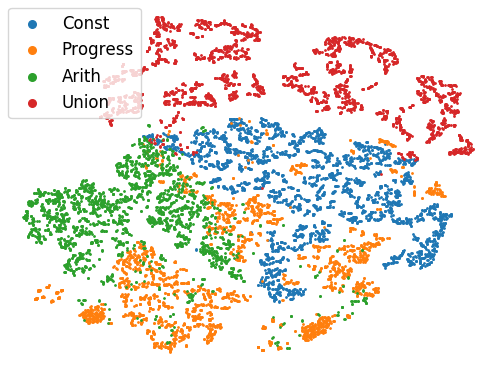}
    \includegraphics[width=0.3\linewidth]{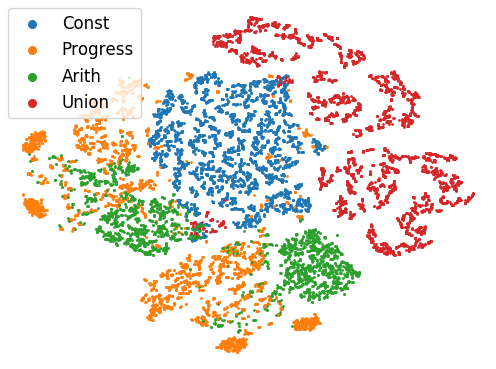}
    \includegraphics[width=0.3\linewidth]{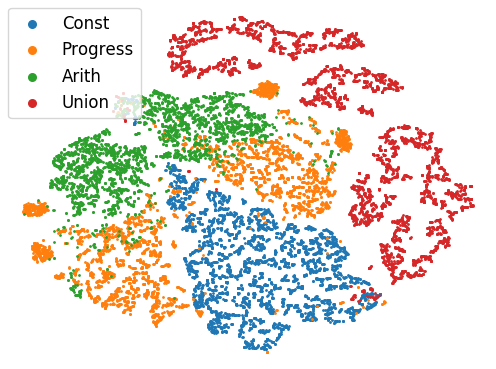}\\
    \hspace{-2em}\texttt{Center}\hspace{10em} \texttt{L-R}\hspace{11em} \texttt{U-D}\\
     \includegraphics[width=0.3\linewidth]{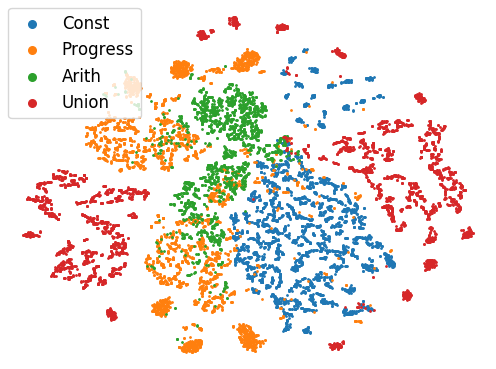}
    \includegraphics[width=0.3\linewidth]{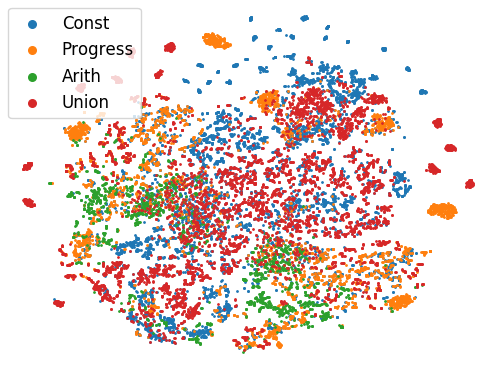}\\ 
    \texttt{O-IC}\hspace{11em} \texttt{O-IG}\\
    \includegraphics[width=0.3\linewidth]{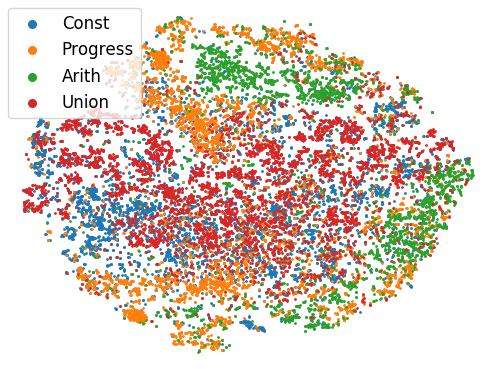}
     \includegraphics[width=0.3\linewidth]{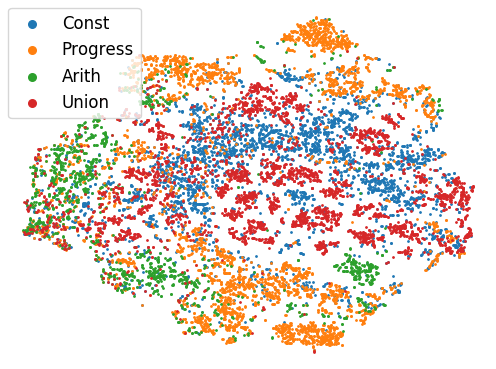}\\
     \texttt{2Grid}\hspace{11em} \texttt{3Grid}
    \caption{t-SNE plots on the output of the relationship neural network $\mathcal{N}^{r}$, in the order: \texttt{Center}, \texttt{L-R}, \texttt{U-D}, \texttt{O-IC}, \texttt{O-IG}, \texttt{2Grid}, \texttt{3Grid}.}
    \label{fig:tsne_all}
\end{figure}

We visualized the relationship learning in all configurations via t-SNE in Fig \ref{fig:tsne_all}, where we observed clusters were well separated in \texttt{Center}, \texttt{L-R}, \texttt{U-D}, and \texttt{O-IC}, but became less distinguishable in the other three tasks. 

\newpage

\section{Further Results on Out-Of-Distribution Generalization}
\subsection{Generalization to a different number of attribute relationship pairs}
\label{sec:num_rels}
In this section, we ask if the model can generalize to a problem that consists of a different number of attribute-relationship pairs. In Balanced-RAVEN, within each configuration, the number of attribute-relationship pairs in every problem is fixed. For example, the configuration \texttt{Center} contains 3 pairs, i.e., one needs to discover 3 underlying patterns to solve the problem. A seemingly simple generalization task to humans but can be difficult for learned models is to generalize to a problem that only consists of 2 attribute-relationship pairs. 

To evaluate this kind of generalization, we created two new datasets for each configuration of \texttt{Center}, \texttt{L-R}, \texttt{U-D}, and \texttt{O-IC}. The first dataset, denoted as ``Rel-1'', defines 1 relationship on each object; the second dataset, denoted as ``Rel-2'', defines 2 relationships on each object. For those attributes that are not associated with any relationships, we randomly sample them. We denote the original dataset, in which there are 3 relationships defined on each object as ``Rel-3''. We performed training and validation on each of the datasets while using the remaining 2 datasets as the test set.
We present the results in Table \ref{tab:diff_rels_tasks_ce}-\ref{tab:diff_rels_tasks_oic}, with comparisons to CoPINet. SCL was able to generalize within $1\%$ accuracy drop from training dataset to test dataset in 22 out of 24 tasks. In contrast, the baseline model LEN severely suffered from a distribution shift from training to test. The accuracy went down from $46.0\%$ to $34.1\%$ when training on ``Rel-3" and testing on ``Rel-1’‘ on \texttt{Center}, whereas SCL only had a slight degradation of $0.7\%$. 


\label{appendix:num_rels}

\begin{table}[ht]
  \caption{Generalization to various number of relationships on \texttt{Center}. }
  \label{tab:diff_rels_tasks_ce}
  \centering
  \begin{tabular}{llllllll}
    \toprule
    &\multicolumn{6}{c}{Test Accuracy}\\
    \cmidrule(r){2-7}
     Train$\backslash$Test & \multicolumn{2}{c}{Rel-1} & \multicolumn{2}{c}{Rel-2} & \multicolumn{2}{c}{Rel-3} \\
          &CoPI &SCL &CoPI &SCL &CoPI &SCL \\
    \midrule
    {Rel-1}    
    &42.7 & 99.9 &   39.8 & 99.9 &    37.8 & 99.7 \\
    {Rel-2}    
    &41.3 & 99.9 &   40.5 & 99.9 &    41.7 & 99.8 \\
    {Rel-3}    
    &34.1 & 99.0 &   41.1 & 99.2 &    46.0 & 99.7 \\
    \bottomrule
  \end{tabular}
  \caption{Generalization to various number of relationships on \texttt{L-R}. }
  \label{tab:diff_rels_tasks_lr}
  \centering
  \begin{tabular}{llllllll}
    \toprule
    &\multicolumn{6}{c}{Test Accuracy}\\
    \cmidrule(r){2-7}
     Train$\backslash$Test & \multicolumn{2}{c}{Rel-1} & \multicolumn{2}{c}{Rel-2} & \multicolumn{2}{c}{Rel-3} \\
         &CoPI &SCL &CoPI &SCL &CoPI &SCL \\
    \midrule
    {Rel-1}    
    &36.8 & 99.9 &   41.0 & 99.9 &    44.8 & 99.9 \\
    {Rel-2}    
    &39.1 & 99.9 &   44.3 & 99.8 &    44.9 & 99.6 \\
    {Rel-3}    
    &35.1 & 99.5 &   43.7 & 99.3 &    47.6 & 99.6 \\
    \bottomrule
  \end{tabular}
  \caption{Generalization to various number of relationships on \texttt{U-D}. }
  \label{tab:diff_rels_tasks_ud}
  \centering
  \begin{tabular}{llllllll}
    \toprule
    &\multicolumn{6}{c}{Test Accuracy}\\
    \cmidrule(r){2-7}
     Train$\backslash$Test & \multicolumn{2}{c}{Rel-1} & \multicolumn{2}{c}{Rel-2} & \multicolumn{2}{c}{Rel-3} \\
         &CoPI &SCL &CoPI &SCL &CoPI &SCL \\
    \midrule
    {Rel-1}    
    &37.3 & 99.9 &   39.7 & 100.0 &   42.1 & 100.0 \\
    {Rel-2}    
    &35.7 & 99.9 &   42.6 & 99.9 &    46.3 & 99.8 \\
    {Rel-3}    
    &34.1 & 99.1 &   42.7 & 99.3 &    46.9 & 99.6 \\
    \bottomrule
  \end{tabular}
  \caption{Generalization to various number of relationships on \texttt{O-IC}. }
  \label{tab:diff_rels_tasks_oic}
  \centering
  \begin{tabular}{llllllll}
    \toprule
    &\multicolumn{6}{c}{Test Accuracy}\\
    \cmidrule(r){2-7}
     Train$\backslash$Test& \multicolumn{2}{c}{Rel-1} & \multicolumn{2}{c}{Rel-2} & \multicolumn{2}{c}{Rel-3} \\
         &CoPI &SCL &CoPI &SCL &CoPI &SCL \\
    \midrule
    {Rel-1}    
    &42.6 & 98.9 &   44.1 & 99.1 &    43.8 & 99.1 \\
    {Rel-2}    
    &37.4 & 99.3 &   43.7 & 99.6 &    45.6 & 99.5 \\
    {Rel-3}    
    &34.5 & 77.9 &   45.4 & 90.4 &    50.0 & 99.2 \\
    \bottomrule
  \end{tabular}
\end{table}

\newpage

\subsection{Generalization to Unseen Analogies in \texttt{L-R}}
\label{appendix:novel_pairs_lr}
In addition to the unseen analogy experiments for the joint-training setting (Section~\ref{sec:ood}), we also performed further experiments on the single training setting of \texttt{L-R}.

Similarly done as in Section~\ref{sec:ood}, we created nine datasets for nine attribute-relationship pairs with 60\%/20\%/20\% train/valid/test split for \texttt{L-R}. We reported both the test accuracy as well as the performance drop from the validation accuracy in Table \ref{tab:att_rel}. We observed that our model could generalize almost perfectly on 8 out of 9 tasks, with an average test accuracy of $97.8\%$. SCL achieves similar validation and test performance with an average gap of $2.2\%$. We compared our model to the existing baselines LEN and CoPINet, and observed large gaps from validation accuracy to test accuracy, an average of $14.1\%$ for CoPINet and $5.3\%$ for LEN, showing their poor generalization to unseen analogies.

\begin{table}[h]
\centering 
\caption{\small Generalization to unseen attribute-relationship pairs.}
  \label{tab:att_rel}
  \begin{adjustbox}{width=\linewidth}
  \begin{tabular}{lllllllllll}
    \toprule
    &\multicolumn{9}{c}{Test Accuracy (Difference to Validation Accuracy)}\\
    \cmidrule(r){2-10}
     & \multicolumn{3}{c}{Type} & \multicolumn{3}{c}{Size} & \multicolumn{3}{c}{Color} \\
     &{\bfseries\textit{LEN}}~\cite{LEN_NIPS2019}
     &{\bfseries\textit{CoPINet}}~\cite{copinet}
     & {\bfseries\textit{SCL}}~(ours) &{\bfseries\textit{LEN}}~\cite{LEN_NIPS2019}
     &{\bfseries\textit{CoPINet}}~\cite{copinet}
     &{\bfseries\textit{SCL}}~(ours) &{\bfseries\textit{LEN}}~\cite{LEN_NIPS2019}
     &{\bfseries\textit{CoPINet}}~\cite{copinet}
     &{\bfseries\textit{SCL}}~(ours) \\
    \midrule
    {Constant}         & 28.0 (-0.8) & 25.1 (-25.9) & \textbf{100.0 (-0)}  & 24.4 (-1.1) & 37.2 (-6.1) & \textbf{99.9 (-0)}  & 25.3 (-1.4) & 38.8 (-4.9) & \textbf{100.0 (-0)} \\
    {Progression}      & 24.0 (-1.3) & 36.2 (-19.2) & \textbf{98.1 (-0.7)}  & 27.9 (-2.5) & 36.2 (-7.9) & \textbf{100.0 (+0.1)}  & 25.3 (-3.2) & 35.8 (-5.9) & \textbf{99.8 (-0.1)} \\
    {Union}            & 29.4 (-11.2) & 32.9 (-24.7) & \textbf{83.3 (-16.4)}  & 27.6 (-12.3) & 36.4 (-18.1) & \textbf{98.8 (-1.1)}  & 22.0 (-13.6) & 29.2 (-14.6) & \textbf{100.0 (-0)} \\

    \bottomrule
  \end{tabular}
  \end{adjustbox}
  \vspace{-1em}
\end{table}

\end{document}